\documentclass[journal]{IEEEtran}
\def\PREPRINT{}

\usepackage{cite}
\usepackage{amsmath,amssymb}
\usepackage{graphicx}
\usepackage{xcolor}
\usepackage{colortbl}
\usepackage{array}
\newcolumntype{L}[1]{>{\raggedright\arraybackslash}p{#1}}

\usepackage{booktabs}
\usepackage{url}
\usepackage{orcidlink}
\usepackage{tikz}
\usepackage{pgfplots}
\pgfplotsset{compat=1.18}
\usetikzlibrary{arrows.meta,positioning,calc,shadows}

\newcommand{\cawhy}[3]{%
  \fill[obnavy!6,rounded corners=3pt] (#1,5.95) rectangle (#1+3.75,8.30);
  \draw[obnavy!40,rounded corners=3pt,line width=0.7pt]
       (#1,5.95) rectangle (#1+3.75,8.30);
  \fill[obamber,rounded corners=2pt] (#1+0.16,7.60) rectangle (#1+1.56,8.16);
  \node[font=\scriptsize\bfseries,text=obnavy] at (#1+0.86,7.88)
       {\scalebox{0.8}{#2}};
  \node[anchor=north west,text width=3.4cm,font=\scriptsize,text=obnavy]
       at (#1+0.18,7.38) {#3};
}

\newcommand{\cacard}[8]{%
  \pgfmathsetmacro{\cy}{#2-0.375}
  \fill[#3!8] ($(#1,\cy)+(-2.1,-1.30)$) rectangle ($(#1,\cy)+(2.1,1.30)$);
  \draw[#3!70,line width=0.7pt] ($(#1,\cy)+(-2.1,-1.30)$) rectangle ($(#1,\cy)+(2.1,2.05)$);
  \fill[#3] ($(#1,\cy)+(-2.1,1.30)$) rectangle ($(#1,\cy)+(2.1,2.05)$);
  \draw[#3!70,line width=0.7pt] ($(#1,\cy)+(-2.1,1.30)$) -- ($(#1,\cy)+(2.1,1.30)$);
  \node[text=#4,font=\small\bfseries] at ($(#1,\cy)+(0,1.67)$) {#5};
  \node[anchor=west,text width=3.75cm,font=\scriptsize,text=obnavy]
     at ($(#1,\cy)+(-1.95,0.82)$) {#6};
  \node[anchor=west,text width=3.75cm,font=\scriptsize,text=obnavy]
     at ($(#1,\cy)+(-1.95,0.0)$) {#7};
  \node[anchor=west,text width=3.75cm,font=\scriptsize,text=obnavy]
     at ($(#1,\cy)+(-1.95,-0.82)$) {#8};
}

\newcommand{\trstage}[7]{%
  \pgfmathsetmacro{\yh}{#3-0.55}
  \fill[#4!8] (#1-1.85,#2) rectangle (#1+1.85,\yh);
  \fill[#4] (#1-1.85,\yh) rectangle (#1+1.85,#3);
  \draw[#4!70,line width=0.7pt] (#1-1.85,#2) rectangle (#1+1.85,#3);
  \draw[#4!70,line width=0.7pt] (#1-1.85,\yh) -- (#1+1.85,\yh);
  \node[font=\small\bfseries,text=#5] at (#1,{(\yh+#3)/2}) {#6};
  \node[text width=3.4cm,align=center,font=\scriptsize,text=obnavy]
       at (#1,{(#2+\yh)/2}) {#7};
}

\definecolor{obnavy}{HTML}{011E4B}
\definecolor{obamber}{HTML}{FAB001}
\definecolor{obsky}{HTML}{87C6E9}
\definecolor{obred}{HTML}{E40613}
\definecolor{obskyd}{HTML}{1B6FA8}
\definecolor{obamberd}{HTML}{A66E00}
\definecolor{obredd}{HTML}{C1050F}
\definecolor{obink}{HTML}{0E1A2B}
\definecolor{obmuted}{HTML}{8493A9}
\definecolor{obgood}{HTML}{178A52}

\newcommand{\hdA}{}
\newcommand{\hdS}{}
\newcommand{\hdR}{}

\begin{document}

\title{OBER+: Continuity-Aware Reporting and Traceable\\
Continuous Improvement in Outcome-Based Education}

\author{Elakkiya~Rajasekar~\orcidlink{0000-0002-2257-0640},~\IEEEmembership{Member,~IEEE}%
\thanks{E. Rajasekar is an Associate Professor with the Department
of Computer Science, Birla Institute of Technology and Science
Pilani, Dubai Campus, Dubai, United Arab Emirates
(e-mail: \texttt{elakkiya@dubai.bits-pilani.ac.in},
ORCID \texttt{0000-0002-2257-0640}).}}

\ifdefined\PREPRINT
  \pagestyle{plain}
\else
  \markboth{IEEE Transactions on Learning Technologies}%
  {Rajasekar: Continuity-Aware Reporting in Outcome-Based Education}
\fi

\maketitle

\begin{abstract}
Institutions practising outcome-based education compute learning
outcome attainment routinely, while reviews of curriculum analytics
report an absence of evidence on how that computation informs decisions
and on what those decisions achieve. This paper presents OBER+, an
extension of a deployed institutional attainment platform that computes
the step from a measured shortfall to an evaluated corrective action.
Five connected stages accumulate attainment across deliveries of a
course, signal a shortfall and a persistent shortfall, grade it on
cutoffs the regulator already uses, record the decision taken against a
catalogue of practices annotated with their evidence, log the resulting
change, and quantify the subsequent movement in the earlier shortfall. A further rule compares successive statements of an outcome,
so that attainment is never read as a series across a point at which
the outcome itself changed. Applying the rules to the live record of
two real courses produced three results. Every outcome of a core course
was substantively redefined between consecutive deliveries, with
subject matter moving between outcome numbers, so a naive reading would
have reported a twenty-five point collapse between quantities that do
not refer to the same learning. Recomputing the platform's figures from
its documented rule showed six of ten differing by more than rounding
explains, in a pattern that identified a defect since reported to the
institution. Across fifteen statement pairs from three transitions, five were
identical character for character, and among the ten that were not, the
outcome carrying a given number was nearest to a differently numbered
earlier outcome in six, a result resting on an ordering of similarities
and requiring no threshold and no labelling. The contribution is a
computational design for outcome-based reporting, stated as rules that
any platform computing attainment per delivery can implement, together
with evidence of what those rules make visible in a live institutional
record.
\end{abstract}


\section{Introduction}
\IEEEPARstart{O}{utcome-based} education organises a course around
statements of what a student should be able to do on completing it.
These statements are the course learning outcomes (CLOs), each linked
to one or more programme learning outcomes (PLOs) that describe what a
graduate of the whole programme should achieve. A CLO is assessed
through the evaluation components of its course, among them quizzes,
assignments, the midsemester examination and the comprehensive
examination. The marks scored in those components are aggregated into a
single figure per outcome, its attainment, which is then compared with
a target percentage.

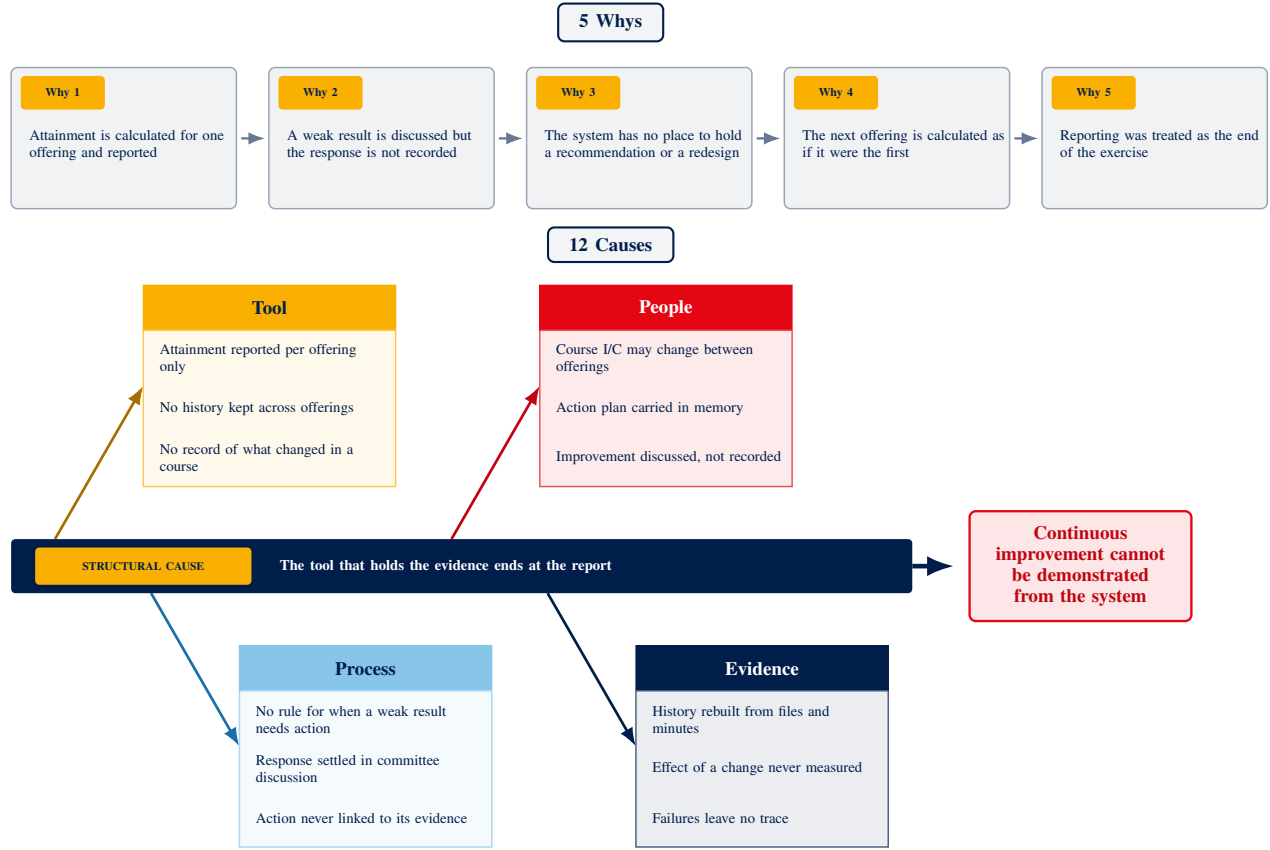
\begin{figure*}[!t]
\centering
\resizebox{0.92\textwidth}{!}{%
\begin{tikzpicture}[>=Latex]

\node[draw=obnavy,line width=1.0pt,rounded corners=3pt,fill=obnavy!5,
      font=\small\bfseries,text=obnavy,inner xsep=10pt,inner ysep=5pt]
      at (9.68,9.05) {5 Whys};
\cawhy{-0.30}{Why 1}{Attainment is calculated for one offering and reported}
\draw[obnavy!60,line width=1.1pt,->] (3.53,7.12) -- (3.93,7.12);
\cawhy{3.99}{Why 2}{A weak result is discussed but the response is not recorded}
\draw[obnavy!60,line width=1.1pt,->] (7.82,7.12) -- (8.22,7.12);
\cawhy{8.28}{Why 3}{The system has no place to hold a recommendation or a redesign}
\draw[obnavy!60,line width=1.1pt,->] (12.11,7.12) -- (12.51,7.12);
\cawhy{12.57}{Why 4}{The next offering is calculated as if it were the first}
\draw[obnavy!60,line width=1.1pt,->] (16.40,7.12) -- (16.80,7.12);
\cawhy{16.86}{Why 5}{Reporting was treated as the end of the exercise}

\node[draw=obnavy,line width=1.0pt,rounded corners=3pt,fill=obnavy!5,
      font=\small\bfseries,text=obnavy,inner xsep=10pt,inner ysep=5pt]
      at (9.68,5.35) {12 Causes};

\draw[obamberd,line width=1.3pt,->] (0.43,0.45)  -- (1.90,3.00);
\draw[obskyd,line width=1.3pt,->]   (2.03,-0.45) -- (3.50,-3.00);
\draw[obredd,line width=1.3pt,->]   (7.03,0.45)  -- (8.50,3.00);
\draw[obnavy,line width=1.3pt,->]   (8.63,-0.45) -- (10.10,-3.00);

\fill[obnavy,rounded corners=2pt] (-0.3,-0.45) rectangle (14.7,0.45);
\draw[obnavy,line width=2.2pt,->] (14.7,0) -- (15.4,0);
\fill[obamber,rounded corners=2pt] (0.10,-0.31) rectangle (3.70,0.31);
\node[font=\scriptsize\bfseries,text=obnavy] at (1.90,0)
  {\scalebox{0.78}{STRUCTURAL CAUSE}};
\node[anchor=west,font=\scriptsize\bfseries,text=white] at (4.05,0)
  {The tool that holds the evidence ends at the report};

\cacard{4.00}{3.00}{obamber}{obnavy}{Tool}
  {Attainment reported per offering only}
  {No history kept across offerings}
  {No record of what changed in a course}
\cacard{10.60}{3.00}{obred}{white}{People}
  {Course I/C may change between offerings}
  {Action plan carried in memory}
  {Improvement discussed, not recorded}
\cacard{5.60}{-3.00}{obsky}{obnavy}{Process}
  {No rule for when a weak result needs action}
  {Response settled in committee discussion}
  {Action never linked to its evidence}
\cacard{12.20}{-3.00}{obnavy}{white}{Evidence}
  {History rebuilt from files and minutes}
  {Effect of a change never measured}
  {Failures leave no trace}

\node[draw=obred,line width=1.1pt,fill=obred!10,rounded corners=3pt,
      text width=3.2cm,align=center,font=\bfseries\small,text=obredd,
      inner sep=7pt] at (17.5,0)
  {Continuous improvement cannot be demonstrated from the system};

\end{tikzpicture}}
\caption{Cause analysis of the improvement gap.}
\label{fig:cause}
\end{figure*}

Computing that figure is a solved problem, automated at institutional
scale by platforms that gather marks, apply the weightings and report
attainment for every outcome of every course
\cite{hussain2021,amirtharaj2022,naim2025}. Acting on it is not.
Criterion~4 of the Accreditation Board for Engineering and Technology
(ABET) requires a documented process for evaluating attainment and for
feeding the results into programme improvement \cite{abet}. The
National Board of Accreditation (NBA) treats revision of outcomes as a
normal corrective action while expecting programme outcomes to remain
stable \cite{nba}. The Commission for Academic Accreditation (CAA) of
the United Arab Emirates requires programmes to define outcomes, assess
them and act on the results, and bands attainment on a four-band scale
\cite{caa}. In published practice a committee discharges that
obligation by reading reports and minutes, so the link between a weak
result, the action taken and the later movement survives only in those
documents \cite{aue,cuboulder}. Fig.~\ref{fig:cause} gives the cause
analysis carried out on the platform studied here, which traced five
successive answers and twelve contributing causes to a single
structural cause, namely that the tool holding the evidence ends at the
report.

Curriculum analytics reaches the same boundary from the other side.
Reviews of the field report that redesigns are made course by course,
and that evidence on how analytics tools inform decisions and affect
outcomes is lacking \cite{drugova2024,desilva2025}. Tools exist that
gather evidence of competency attainment and that screen courses for
review \cite{hilliger2022,joseph}, but in every case the decision and
its consequence lie outside the tool, as Table~\ref{tab:gapmap} sets
out system by system.

Underneath that boundary lies a question the literature does not ask.
Reading attainment across deliveries of a course assumes that the
outcome being measured stayed the same, and nothing in current practice
tests the assumption. Outcome statements are revised between
deliveries, most often when a course changes hands, and the revision
leaves no trace in a record that goes on reporting figures against the
same outcome numbers. Where the assumption fails, every later step
fails with it, because a trend, a corrective action and any
movement attributed to it all rest on comparing an outcome with itself.

OBER+ is a design that computes the step from a measured shortfall to
an evaluated corrective action, and that tests the assumption as part
of doing so. It extends a deployed institutional attainment platform
\cite{ober} with five connected stages, named Report, Reflect,
Recommend, Redesign and Reassess and referred to together as the 5R
cycle, which run on the attainment the platform already computes.
Stating the design as computational rules rather than as a workflow
allows it to be implemented on any system that reports attainment for
each delivery of a course. One rule compares successive statements of
an outcome by sentence embedding and by the cognitive level of its
leading verb, and classifies what changed.

Three research questions organise the work.

\begin{enumerate}
\item What computational rules turn a record of attainment across
deliveries of a course into an improvement record in which a corrective
action is bound to the evidence that prompted it and to the movement
that follows it?
\item Can it be established automatically, and without labelled data,
whether an attainment series spanning two deliveries refers to the same
learning outcome?
\item What does applying the rules to a live institutional record make
visible that reporting each delivery separately does not?
\end{enumerate}

A design and a public implementation \cite{oberplus} answer the first.
The second and third are answered on a sample of the live record of the
institutional platform, two courses in one department for which the
author is the course in-charge, one of them with two consecutive
deliveries recorded.

The work contributes three things. The first is a computational
traceability model that connects attainment evidence, the corrective
decision taken against it, the change implemented and the movement
observed afterwards. The second is an outcome continuity check that
tests, without a threshold and without labelled data, whether an
attainment series refers to comparable learning before that series is
read as a trend. The third is an application to a live institutional
record which exposes two failure modes that current practice leaves
invisible, namely subject matter moving between outcome numbers, and
disagreement between a platform's documented and implemented attainment
computation. Two of the three results were not anticipated when the
design was written. Every outcome of the two-delivery course was
substantively redefined between deliveries while attainment continued
to be reported against unchanged outcome numbers. The platform's
published figures disagree with its own documented computation on most
outcomes, in a pattern that identifies the cause. Across three
transitions the outcome carrying a given number is frequently nearest
to a differently numbered outcome of the previous delivery, so the
numbering by which an attainment series is indexed does not track the
subject matter that series reports.

Section~\ref{sec:related} places the work against published systems and
against the curriculum analytics literature.
Sections~\ref{sec:design} and~\ref{sec:impl} give the design and its
implementation. Section~\ref{sec:evaldesign} describes the live record
and the claims each part of the evaluation can support,
Section~\ref{sec:findings} reports the results, and
Section~\ref{sec:discussion} discusses their reach and their limits.

\section{Related Work}
\label{sec:related}

\begin{table*}[!t]
\caption{Published continuous improvement systems compared with OBER+.}
\label{tab:gapmap}
\centering
\scriptsize
\renewcommand{\arraystretch}{1.12}
\begin{tabular}{L{32mm}|L{23mm}|L{23mm}|L{25mm}|L{24mm}|L{24mm}}
\hline
\hdA\textbf{System} &
\hdS\textbf{Attainment history} &
\hdR\textbf{Automatic detection} &
\hdA\textbf{Action bound to evidence} &
\hdS\textbf{Post-action movement tracked} &
\hdR\textbf{Outcome revision classified} \\
\hline
Institutional platform, United Arab Emirates~\cite{aue} &
Yes, across cycles & Not reported &
Actions stored, link not reported & No, manual annual meeting & No \\
\hline
Attainment computation systems~\cite{hussain2021,amirtharaj2022,naim2025} &
Per offering only & No & No & No & No \\
\hline
Curriculum analytics tools~\cite{hilliger2022,desilva2025} &
Yes, evidence collection & Not at CLO level &
No, actions outside the tool & No & No \\
\hline
Departmental improvement policy, United States~\cite{cuboulder} &
Committee records & No & No, acting on advice untracked & No & No \\
\hline
Course-level gap screening~\cite{joseph} &
Yes, across offerings & Course level only & No & No & No \\
\hline
Taxonomy-level classification~\cite{li2022,gani2023,almatrafi2025,kumar} &
No, one-time analysis & No & No & No & No, single statement only \\
\hline
OBER+ (this work) & Yes, per CLO per offering &
Yes, alert per offering and flag with severity band &
Yes, by recorded decision & Yes, gap closure per
intervention &
Yes, paraphrase, level change or replacement \\
\hline
\end{tabular}
\end{table*}

\subsection{Outcome-Based Education and Continuous Improvement}

Outcome-based education designs a curriculum from its intended results,
defining the outcomes first and then the teaching and the assessment
that lead to them. The approach was formalised in the early 1990s
\cite{spady} and adopted across professional education \cite{harden}. A review of the field found that although the
approach spread widely, direct empirical evidence for its effectiveness
remained limited \cite{morcke}. That finding bears on the present work,
since outcome-based practice requires evidence of what follows the
actions its assessment results prompt. Two principles carry into the design.
Constructive alignment requires outcomes, teaching and assessment to be
designed as one system \cite{biggs96,biggstang}. Taxonomies supply the
action verbs with which outcomes are written and levelled
\cite{bloom}, later revised into a two-dimensional form
\cite{anderson,krathwohl02}. The move of accreditation to measured
outcomes, its rationale and its effect on programmes are documented in
\cite{prados,besterfield}.

\subsection{Attainment Computation Systems}

Recent systems automate the calculation and stop at the reported
number. A digital framework for programme impact evaluation uses
granular student outcome data for accreditation reporting
\cite{hussain2021}. A web-based system at an autonomous engineering
college computes course, programme and educational-objective attainment
from direct and indirect assessment \cite{amirtharaj2022}, and an
information-system model computes programme outcome attainment from
student scores \cite{naim2025}. Natural language processing has been
applied to align course outcomes to programme outcomes automatically,
so that the computation rests on accurate mappings \cite{zaki2023}. None of these keeps the attainment of an outcome
across deliveries, none raises a signal when an outcome falls or stays
below target, none records the decision taken in response, and none
tracks the movement that follows the resulting change.

Where a system does hold actions, the evaluation of those actions is
still human. An institutional platform at a university in the United
Arab Emirates stores outcome, action and key performance indicator data
across assessment cycles, and evaluates the effectiveness of the
actions in a manual annual meeting \cite{aue}. A published continuous
improvement policy at a United States engineering department assigns
recommendations to an annual committee review and states that no formal
mechanism tracks whether an instructor acted on a recommendation or
whether the action worked \cite{cuboulder}.

\subsection{Curriculum and Learning Analytics}

The analytics literature reaches further and stops at the same
boundary. Course screening by gap analysis and machine learning
identifies which course needs attention without attributing the gap to
any assessment component and without following any intervention that
results \cite{joseph}. Curriculum analytics tools help teachers gather
richer evidence of competency attainment for programme improvement
\cite{hilliger2022}. A review of learning analytics in learning design
finds redesigns made course by course, with no standard way of
attributing a later gain to a specific change \cite{drugova2024}. A
systematic review of curriculum analytics states that evidence on how
such tools inform decisions and affect outcomes is lacking
\cite{desilva2025}. The design presented here responds to that finding.
Binding a recorded decision to the evidence that prompted it, and
computing the movement that followed, is what turns an analytics tool
into a source of such evidence.

\subsection{Classification of Outcome Statements}

Automatic classification of learning outcomes onto taxonomy levels has
reached practical accuracy. Transformer models classify
learning objectives at scale on a corpus of more than twenty thousand
objectives \cite{li2022}, and convolutional networks with pre-trained
embeddings classify examination questions \cite{gani2023}. Generative
language models now match or exceed those classifiers on course
learning outcomes \cite{almatrafi2025}, and the task has been addressed
for outcomes and examination questions together \cite{kumar}. All of this
work classifies a statement once, at a single point in time. No
published application repeats the classification across deliveries of a
course in order to detect that the wording of an outcome changed.
Sentence embeddings compared by cosine similarity are the established
means of judging whether two statements express the same thing and are
trained and evaluated on that task directly \cite{reimers}. The rule
given in Section~\ref{sec:design} combines that measure with the
taxonomy level, so that a revision is not only detected but classified.

\subsection{Evidence Base for Improvement Practices}

The premise of the improvement cycle is that assessment information
raises achievement when it reaches a person who can act on it in a
usable form. The formative assessment literature establishes that
\cite{blackwiliam,nicol,hattie}, and a recent systematic review
confirms it for higher education while noting uneven evidence across
disciplines \cite{morris2021}. That literature is written for the
classroom and the programme review, and none of it addresses how an
institution's own attainment platform should deliver the information to
the person responsible for a course.

The practices a platform can offer differ in how well they are
supported, and the design recorded here grades them rather than
presenting a flat list. Active learning, peer instruction, the flipped
classroom, supplemental instruction and retrieval practice each carry
meta-analytic or systematic review evidence
\cite{freeman,theobald,prince,xu2026,crouch,oz2024,lohew,bredow2021,dawson2014,roediger,dunlosky,yang2021},
and worked examples and assessment redesign rest on cognitive load
theory \cite{sweller,paas}. Two options carry weaker
evidence and are labelled as such. Industry-sourced assessment content
rests on an authentic assessment blueprint with no quantified
attainment effect \cite{villarroel}, and remediation assisted by
artificial intelligence reports feasibility and acceptance rather than
measured learning \cite{alfoori}. How that grading enters the design is described in
Section~\ref{sec:design}.

Table~\ref{tab:gapmap} places the reviewed systems against the
capabilities designed here. No reviewed system keeps an attainment record per outcome across
deliveries, detects a shortfall automatically at outcome level, binds a
corrective action to the evidence that prompted it, or tracks the
movement that follows, and none classifies a revision of an outcome
statement across deliveries. The comparison covers the systems reviewed
above, selected as those that compute or report outcome attainment at
institutional scale, those that store corrective actions, and those
that classify outcome statements. It is not the product of a systematic
search, so the claim it supports concerns the reviewed systems rather
than the whole published literature.

\section{Framework Design}
\label{sec:design}

\subsection{Design Principles}

The design is governed by three principles. It extends a deployed
platform rather than replacing it, so every operational 5R rule reads
the attainment the platform has already computed and no such rule
recomputes or reweights it. It computes everything that can be computed and leaves exactly one
thing to the person responsible for the course, called the course
in-charge (course I/C) on the campus studied here, which is the
improvement decision itself. Every record it creates carries the
identifiers needed to connect it to the records before and after it, so
that the path from evidence to recorded movement is never broken.

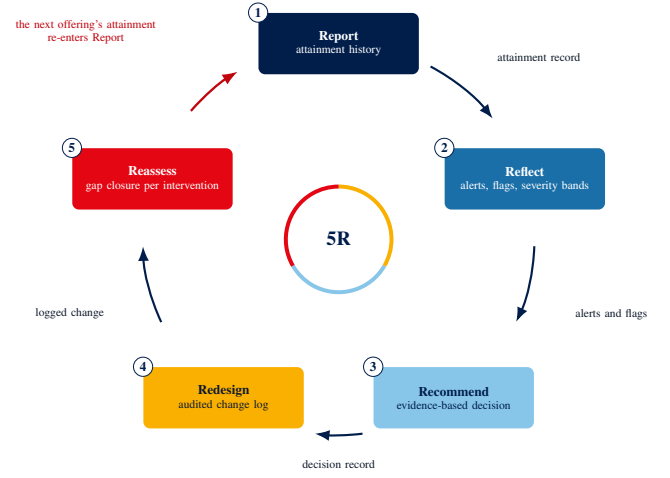
\begin{figure}[!t]
\centering
\resizebox{0.96\columnwidth}{!}{%
\begin{tikzpicture}[
  num/.style={circle, fill=white, draw=obnavy, line width=0.6pt,
    text=obnavy, font=\bfseries\footnotesize, inner sep=1.7pt},
  arr/.style={-{Latex}, line width=1.3pt, color=obnavy},
  ret/.style={-{Latex}, line width=1.4pt, color=obredd}]
  \def\obR{4.15}
  \node[rectangle, rounded corners=4pt, fill=obnavy, text=white,
    font=\footnotesize, minimum width=34mm, minimum height=13mm,
    align=center] (r1) at (90:\obR)
    {\textbf{Report}\\[-1pt]{\scriptsize attainment history}};
  \node[rectangle, rounded corners=4pt, fill=obskyd, text=white,
    font=\footnotesize, minimum width=34mm, minimum height=13mm,
    align=center] (r2) at (18:\obR)
    {\textbf{Reflect}\\[-1pt]{\scriptsize alerts, flags, severity bands}};
  \node[rectangle, rounded corners=4pt, fill=obsky, text=obnavy,
    font=\footnotesize, minimum width=34mm, minimum height=13mm,
    align=center] (r3) at (-54:\obR)
    {\textbf{Recommend}\\[-1pt]{\scriptsize evidence-based decision}};
  \node[rectangle, rounded corners=4pt, fill=obamber, text=obink,
    font=\footnotesize, minimum width=34mm, minimum height=13mm,
    align=center] (r4) at (-126:\obR)
    {\textbf{Redesign}\\[-1pt]{\scriptsize audited change log}};
  \node[rectangle, rounded corners=4pt, fill=obred, text=white,
    font=\footnotesize, minimum width=34mm, minimum height=13mm,
    align=center] (r5) at (162:\obR)
    {\textbf{Reassess}\\[-1pt]{\scriptsize gap closure per intervention}};
  \node[num] at (r1.north west) {1};
  \node[num] at (r2.north west) {2};
  \node[num] at (r3.north west) {3};
  \node[num] at (r4.north west) {4};
  \node[num] at (r5.north west) {5};
  \draw[arr] (62:\obR)   arc (62:38:\obR);
  \draw[arr] (-2:\obR)   arc (-2:-25:\obR);
  \draw[arr] (-83:\obR)  arc (-83:-97:\obR);
  \draw[arr] (-155:\obR) arc (-155:-178:\obR);
  \draw[ret] (139:\obR)  arc (139:121:\obR);
  \node[font=\scriptsize, color=obink, anchor=west] at (50:5.05)
    {attainment record};
  \node[font=\scriptsize, color=obink, anchor=west] at (-18:5.15)
    {alerts and flags};
  \node[font=\scriptsize, color=obink, anchor=north] at (-90:4.55)
    {decision record};
  \node[font=\scriptsize, color=obink, anchor=east] at (-162:5.1)
    {logged change};
  \node[font=\scriptsize, color=obredd, align=center, anchor=south east]
    at (133:5.5)
    {the next offering's attainment\\ re-enters Report};
  \draw[line width=2.4pt, color=obamber] (90:1.12) arc (90:-30:1.12);
  \draw[line width=2.4pt, color=obsky]   (-30:1.12) arc (-30:-150:1.12);
  \draw[line width=2.4pt, color=obred]   (-150:1.12) arc (-150:-270:1.12);
  \node[font=\large\bfseries, color=obnavy] at (0,0.02) {5R};
\end{tikzpicture}}
\caption{The 5R continuous improvement cycle.}
\label{fig:loop}
\end{figure}

They follow from the cause analysis of Fig.~\ref{fig:cause}, since a
structural cause calls for an extension of the tool rather than a
process placed beside it. The five stages are shown in
Fig.~\ref{fig:loop}. Report keeps the attainment record, Reflect
derives the signals from it, Recommend records the decision, Redesign
logs the change, and Reassess computes the gap closure that follows the
change. The
subsections below are organised by what the rules do rather than by
stage name, and each names the stage it belongs to.

\subsection{Attainment Model and Notation}

One delivery of a course in one semester is called an offering.
Notation is fixed here for one course. $S$ is the set of enrolled
students, $K$ the set of evaluation components, $C$ the set of CLOs and
$J$ the set of PLOs. The mark distribution matrix entry $M_{c,k}$ is
the number of marks component $k$ allocates to CLO $c$. The weightage
distribution matrix entry $w_{c,k}$ is the weight in percent that
component $k$ carries in CLO $c$, and every CLO satisfies
$\sum_{k \in K} w_{c,k} = 100$. The marks student $s$ scored on the questions of
component $k$ that assess CLO $c$ are $m_{s,c,k}$, bounded by $0 \le
m_{s,c,k} \le M_{c,k}$. Attainment is a chain of four means. The class
mean for one CLO within one component is
\begin{equation}
a_{c,k} \;=\; \frac{\sum_{s \in S} m_{s,c,k}}{|S| \, M_{c,k}} ,
\label{eq:component}
\end{equation}
the CLO combines its components over its full weightage row,
\begin{equation}
A_c \;=\; \frac{1}{100} \sum_{k \in K} w_{c,k} \, a_{c,k} ,
\label{eq:clo}
\end{equation}
the course attainment of (\ref{eq:course}) weights each CLO by its mark share
$\mu_c = \sum_{k \in K} M_{c,k}$,
\begin{equation}
A_{\text{course}} \;=\;
\frac{\sum_{c \in C} \mu_c \, A_c}{\sum_{c \in C} \mu_c} ,
\label{eq:course}
\end{equation}
and each PLO is the plain mean of the CLOs mapped to it, with
$\pi_{c,j} = 1$ when CLO $c$ is mapped to PLO $j$ and $0$ otherwise,
\begin{equation}
P_j \;=\;
\frac{\sum_{c \in C} \pi_{c,j} \, A_c}{\sum_{c \in C} \pi_{c,j}} ,
\qquad \pi_{c,j} \in \{0,1\} .
\label{eq:plo}
\end{equation}
Keeping the full weightage row in the denominator of (\ref{eq:clo})
means that a component whose marks have not yet been uploaded
contributes zero, so a partially assessed CLO is never overstated. As a
concrete instance, the live First Semester 2025--26 offering of
CS~F351 Theory of Computation on the campus studied here had 238
enrolled students and six evaluation components carrying 200 marks in
total. The quiz allocates 30 marks to CLO1 and the class mean on those
questions is 18.07, so $a_{\mathrm{CLO1,Quiz}} = 0.602$, and with
weights 60, 20 and 20 over the quiz, the midsemester examination and
the comprehensive examination, (\ref{eq:clo}) gives
$A_{\mathrm{CLO1}} = 0.509$.

Writing $A_c^{(t)}$ for the attainment of item $c$ in offering $t$,
read as a percentage, and $T_c$ for its target, everything that follows
reads $A_c^{(t)}$ and nothing else. This is the interface across which
the design transfers, since any platform that reports attainment per
offering supplies it. The target defaults to 60 percent and is editable
per course, since neither the platform nor the accreditation frameworks
fix a course-level target. That value is the threshold at which the CAA
assessment quality indicator enters its Medium band \cite{caa}.

\subsection{Minimum History Requirement}

Report accumulates the attainment of every CLO, PLO and course into a
record stored per course, per item and per offering. The record is
counted in offerings rather than calendar years, because a course may
be offered once or twice a year and the evidence lies in deliveries
rather than dates. No trend judgement is made until three offerings
exist. With $H_c$ denoting the set of offerings in which item $c$ has a
recorded attainment, the rule is
\begin{equation}
\text{gate}(c) \;=\; \bigl[\, |H_c| \;\geq\; 3 \,\bigr] ,
\label{eq:gate}
\end{equation}
and only when it holds are the last three offerings, written
$W = \{t-2,\, t-1,\, t\}$, read for a trend. Neither ABET, NBA nor CAA states how many
offerings of data justify reading a pattern as a trend, and neither do
the reviewed systems, so this threshold is an explicit design decision
rather than a requirement inherited from elsewhere. It governs
the trend judgement alone and does not delay action, for the reason
given next.

\subsection{Shortfall Detection and Severity Banding}

Reflect turns the record into signals at two levels. An item below
target in the current offering raises an alert,
\begin{equation}
\text{alert}(c,t) \;=\; \bigl[\, A_c^{(t)} < T_c \,\bigr] ,
\label{eq:alert}
\end{equation}
so that the course I/C sees the shortfall in the offering in which it
occurred and can record a decision on it at once. The second level is
the trend judgement and requires the three offerings of
(\ref{eq:gate}). An item is flagged when it misses its target in at
least two of the last three offerings,
\begin{equation}
\text{flag}(c) \;=\;
\Bigl[\, \bigl|\{\, \tau \in W : A_c^{(\tau)} < T_c \,\}\bigr|
\;\geq\; 2 \,\Bigr] .
\label{eq:flag}
\end{equation}
Two levels are needed because a single mechanism cannot serve both
purposes. A rule strict enough to justify the claim that a weakness
persists is too slow to help the course I/C teaching now, and a rule
fast enough to help immediately cannot support that claim. One weak
offering therefore raises an alert but not a flag, and one good
offering does not clear a real decline.

A flag alone does not say how bad the shortfall is. For a flagged item
the severity is the average shortfall over the offerings that missed,
\begin{equation}
\sigma_c \;=\;
\frac{1}{|W_c^-|} \sum_{\tau \in W_c^-}
\bigl(T_c - A_c^{(\tau)}\bigr) ,
\label{eq:severity}
\end{equation}
taken over $W_c^- = \{\, \tau \in W : A_c^{(\tau)} < T_c \,\}$ and
expressed in percentage points, and the band is assigned on the
attainment-to-target ratio
\begin{equation}
\rho_c \;=\; \frac{100}{|W_c^-|}
\sum_{\tau \in W_c^-} \frac{A_c^{(\tau)}}{T_c} ,
\label{eq:rho}
\end{equation}
by the four-band rule
\begin{equation}
\text{band}(\rho_c) =
\begin{cases}
\text{High} & \rho_c \geq 90 \\
\text{Medium} & 60 \leq \rho_c < 90 \\
\text{Low} & 30 \leq \rho_c < 60 \\
\text{Very Low} & \rho_c < 30 .
\end{cases}
\label{eq:band}
\end{equation}
The cutoff spacing is that of the CAA rubric \cite{caa}, so the
severity vocabulary is one the regulator already uses rather than a new
scale a reviewer would have to accept. An item that is not flagged has
no shortfall to band and is reported as on target.

\subsection{Outcome Revision Detection}

Attainment can be compared from one offering to the next only while the
outcome being measured remains the same. Revision of an outcome
statement between offerings is common when a course changes hands, and
it is invisible in current practice, so an attainment series can span
offerings that no longer measure the same thing. Reflect therefore
compares the statement of each CLO across the offerings in $W$.

Each statement is represented as a sentence embedding. The embedding
is the mean of the pretrained 300-dimensional word vectors of the
\texttt{en\_core\_web\_md} model of the spaCy library \cite{spacy},
which are distributional word vectors of the kind established for this
purpose \cite{mikolov}. The statement recorded in an offering is
compared with the statement recorded in the offering before it by
cosine similarity,
\begin{equation}
s_t \;=\; \frac{\mathbf{e}_t \cdot \mathbf{e}_{t-1}}
{\lVert \mathbf{e}_t \rVert \, \lVert \mathbf{e}_{t-1} \rVert} ,
\label{eq:sim}
\end{equation}
where $\mathbf{e}_t$ is the embedding of the statement as it stood in
offering $t$. Similarity alone cannot separate a rewording from a
change of demand, because raising an outcome from applying to designing
alters very few words. The leading action verb of each statement is
therefore classified by a lookup on that verb against the Revised Bloom's
Taxonomy (RBT) action-verb list circulated on the campus, which follows
\cite{anderson}. That list assigns several verbs to more than one
level, so the lookup returns a set of levels $B_t$ rather than a single
level, and a change of level is asserted only when the two sets are
disjoint. The lookup is a deliberate simplification whose failure modes
on real handouts are reported in Section~\ref{sec:findings}. Writing $d_t$ for the
statement itself, the revision recorded at offering $t$ is classified
as
\begin{equation}
\text{Revision}_t =
\begin{cases}
\text{Unchanged} & d_t = d_{t-1}\\[2pt]
\text{Replacement} & s_t < \tau\\[2pt]
\text{Level change} & s_t \ge \tau,\; B_t \cap B_{t-1} = \emptyset\\[2pt]
\text{Paraphrase} & \text{otherwise},
\end{cases}
\label{eq:revision}
\end{equation}
where the first case compares the statements up to case and spacing and
$\tau = 0.90$ is a threshold whose setting is reported in
Section~\ref{sec:evaldesign}. Static word vectors and an explicit
threshold are used in preference to a contextual encoder. A
classification that enters an accreditation record has to be
reproducible by an external reviewer and open to inspection, and it has
to run within the resources of the deployed application. The
classification annotates the record and never suppresses a flag, so no
revision of wording can conceal a shortfall. Where the embedding model
is unavailable the check reverts to the comparison of taxonomy levels
alone.

Comparing only the statements that share an outcome number assumes that
the numbering is stable between deliveries, and that assumption can
fail even when every statement is a defensible revision of something.
A second check therefore tests it directly. For outcome $j$ of the
current delivery, the nearest earlier statement is
\begin{equation}
n(j) \;=\; \arg\max_{i \in C_{t-1}}
\frac{\mathbf{e}_{t,j} \cdot \mathbf{e}_{t-1,i}}
{\lVert \mathbf{e}_{t,j} \rVert \, \lVert \mathbf{e}_{t-1,i} \rVert} ,
\label{eq:align}
\end{equation}
and the numbering is reported as aligned for that outcome when
$n(j) = j$. Where it is not, the outcome now carrying number $j$ is
closer to a differently numbered outcome of the previous delivery, so
an attainment series indexed by outcome number does not follow the
same subject matter. This check requires no threshold and no taxonomy
level, only the ordering of similarities, so it is independent of both
parameters the classification rule depends on.

\subsection{Decision and Change Records}

Recommend is the one stage that computes nothing and records
everything. For any item that is alerted or flagged it places the
evidence, namely the record, the shortfall and, for a flagged item, the
band, beside a catalogue of improvement practices in three categories.
The standard practices carry the meta-analytic and systematic review
evidence of Section~\ref{sec:related}. The innovative practices carry
weaker or emerging evidence and are labelled as such. The third
category is open ended, so that when neither list fits the situation
the course I/C formulates the action in their own words and it is
recorded on the same terms. Every decision is stored as
\begin{equation}
r \;=\; \bigl(\, \mathrm{id},\; c,\; a,\; \kappa,\; e,\;
u,\; d \,\bigr) ,
\label{eq:decision}
\end{equation}
with category $\kappa \in \{\text{standard},\, \text{innovative},\,
\text{open}\}$, where $c$ is the alerted or flagged item, $a$ is the
action with its supporting citation where one exists, $e$ is the
supporting evidence, $u$ is the deciding course I/C and $d$ is the
date. The catalogue grades its evidence rather than presenting a flat list.
A catalogue that gives a meta-analysis and a feasibility study equal
standing invites the weaker option to be adopted on the strength of the
platform's endorsement.

Redesign maintains a single log for everything that changes in a
course, namely its CLOs, its evaluation components and its mappings. An
entry enters the log in one of two ways. A formal entry is created when
a recorded decision is implemented and carries that decision's
identifier. A detected entry is created by the rule of
(\ref{eq:revision}) when it reports a revision with no recorded
decision behind it. Every entry is
\begin{equation}
\ell \;=\; \bigl(\, \mathrm{id},\; c,\; \chi,\;
\mathrm{before},\; \mathrm{after},\; u,\; d,\; q \,\bigr) ,
\label{eq:logentry}
\end{equation}
of kind $\chi \in \{\text{formal},\, \text{detected}\}$, where before
and after hold the full content of the item on either side of the
change, $u$ and $d$ record the person and the date, and $q$ carries the
decision identifier for a formal entry and the classification for a
detected one. Recording both kinds matters because informal change is
precisely the change that current practice loses.

\subsection{Gap Closure}

Reassess evaluates each logged change between the offerings
immediately before and after it, which means a change is judged on one
offering of evidence rather than on a fresh three-offering record.
With the shortfall defined as $g^{(\tau)} = \max\bigl(0,\, T_c -
A_c^{(\tau)}\bigr)$, gap closure is
\begin{equation}
\text{Closure} \;=\;
\frac{g^{(\text{before})} - g^{(\text{after})}}{g^{(\text{before})}} ,
\label{eq:closure}
\end{equation}
so that a value of 1 means the gap closed fully, 0 means no change and
a negative value means regression. Closure is banded on the cutoffs of
(\ref{eq:band}) read as a percentage, with the bands named Strong at 90
or above, Partial from 60, Limited from 30 and Very limited below 30,
including any negative value. The result is attached to the redesign
record whether it is a success or a failure, and when the next
offering's attainment arrives it enters the record and the cycle
continues. Closure is deliberately a movement measure and not a causal
estimate, and Section~\ref{sec:discussion} states the limits of what it
can be read to mean.

\section{System Implementation}
\label{sec:impl}

The rules are implemented in Python as a single web application built
on the Streamlit framework, version-controlled in the public repository
\texttt{github.com/Elakkiya16/OBER\_Plus} and deployed at
\texttt{oberplus.streamlit.app}~\cite{oberplus}. Every quantity in
Section~\ref{sec:findings} is therefore independently reproducible,
either by opening the deployed application or by running the source.

\begin{figure*}[!t]
\centering
\resizebox{0.80\textwidth}{!}{%
\begin{tikzpicture}[
  shdw/.style={drop shadow={shadow xshift=0.5mm, shadow yshift=-0.5mm,
    opacity=0.18, fill=obnavy}},
  scr/.style={rectangle, rounded corners=4pt, fill=obsky!40,
    draw=obskyd, line width=0.7pt, text=obnavy,
    minimum height=14mm, minimum width=68mm, align=center, shdw},
  logic/.style={rectangle, rounded corners=4pt, fill=obnavy,
    text=white, minimum height=14mm, minimum width=62mm,
    align=center, shdw},
  ent/.style={rectangle, rounded corners=3pt, fill=white,
    draw=obamberd, line width=0.6pt, text=obink,
    font=\scriptsize, minimum height=7.2mm, minimum width=41mm,
    align=center},
  tag/.style={rectangle, rounded corners=5pt, font=\bfseries\tiny,
    text=white, minimum height=5.5mm, inner xsep=2.4mm, shdw},
  badge/.style={circle, fill=white, draw=obskyd, line width=0.6pt,
    inner sep=1.4pt, font=\bfseries\tiny, text=obnavy},
  lbl/.style={font=\scriptsize, color=obink},
  arr/.style={-{latex}, line width=1.3pt, color=obskyd},
  arr2/.style={{latex}-{latex}, line width=1.3pt, color=obskyd}]

  \fill[rounded corners=4pt, obsky!13]   (-7.7,0.0)   rectangle (7.7,-2.45);
  \fill[rounded corners=4pt, obnavy!7]   (-7.7,-2.95) rectangle (7.7,-5.5);
  \fill[rounded corners=4pt, obamber!13] (-7.7,-6.0)  rectangle (7.7,-9.7);
  \node[tag, fill=obskyd,  anchor=west] at (-7.5,0.0)   {INTERFACE};
  \node[tag, fill=obnavy,  anchor=west] at (-7.5,-2.95) {LOGIC};
  \node[tag, fill=obamberd, anchor=west] at (-7.5,-6.0) {DATA};

  \foreach \x/\nm in {-3.6/{course I/C}, 0/{DCA}, 3.6/{CAA core committee}}{
    \fill[obnavy] (\x,3.02) circle (0.11);
    \fill[obnavy] (\x,2.68) arc (0:180:0.18) -- cycle;
    \fill[obnavy] (\x-0.18,2.68) rectangle (\x+0.18,2.62);
    \node[font=\scriptsize\bfseries, color=obnavy, anchor=north]
      at (\x,2.52) {\nm};}
  \node[rectangle, rounded corners=4pt, fill=obred!10, draw=obred,
    line width=1.1pt, text=obredd, font=\bfseries\small,
    minimum height=9.5mm, minimum width=142mm, align=center, shdw]
    (shell) at (0,0.95) {OBER+ application shell and navigation};

  \node[scr] (ober) at (-3.6,-1.35)
    {\textbf{OBER screens}\\[1pt]{\scriptsize CLO to PLO mapping · assessment · reports}};
  \node[badge, minimum size=4.6mm] at (-6.35,-0.77) {};
  \draw[obnavy, line width=0.5pt] (-6.475,-0.71) rectangle (-6.225,-0.85);
  \draw[obnavy, line width=0.5pt] (-6.475,-0.75) -- (-6.225,-0.75);
  \node[scr] (plus) at (3.6,-1.35)
    {\textbf{5R screens}\\[1pt]{\scriptsize Report\,·\,Reflect\,·\,Recommend\,·\,Redesign\,·\,Reassess}};
  \node[badge, minimum size=4.6mm] at (0.85,-0.77) {\tiny 5R};

  \node[logic] (comp) at (-3.6,-4.22)
    {\textbf{Attainment computation}\\[1pt]{\scriptsize marks $\to$ component $\to$ CLO $\to$ course $\to$ PLO}};
  \node[circle, fill=white, inner sep=1.4pt, font=\bfseries\tiny,
    text=obnavy] at (-6.35,-3.64) {$\Sigma$};
  \node[logic] (eng) at (3.6,-4.22)
    {\textbf{5R engine}\\[1pt]{\scriptsize alert · flag · severity · band · wording · closure}};
  \foreach \a in {0,45,...,315}{
    \fill[white, rotate around={\a:(0.85,-3.64)}]
      (0.77,-3.83) rectangle (0.93,-3.78);}
  \fill[obnavy] (0.85,-3.64) circle (0.115);
  \draw[white, line width=0.6pt] (0.85,-3.64) circle (0.115);
  \fill[white] (0.85,-3.64) circle (0.035);

  \fill[obamber!28] (-6.9,-6.85) -- (-6.9,-9.15)
    arc (180:360:6.9 and 0.32) -- (6.9,-6.85)
    arc (0:180:6.9 and -0.32) -- cycle;
  \fill[obamber!45] (6.9,-6.85) arc (0:360:6.9 and 0.32);
  \draw[obamberd, line width=0.7pt] (-6.9,-6.85) -- (-6.9,-9.15)
    arc (180:360:6.9 and 0.32) -- (6.9,-6.85);
  \draw[obamberd, line width=0.7pt] (6.9,-6.85) arc (0:360:6.9 and 0.32);
  \node[font=\scriptsize\bfseries, color=obink] at (0,-6.85) {data store};
  \node[ent] at (-4.7,-7.7) {courses and CLOs};
  \node[ent] at (0,-7.7) {weightage and mark matrices};
  \node[ent] at (4.7,-7.7) {marks per offering};
  \node[ent] at (-4.7,-8.62) {CLO to PLO mapping};
  \node[ent] at (0,-8.62) {R3 decision records};
  \node[ent] at (4.7,-8.62) {R4 change log};

  \draw[arr2] (0,2.1) -- (shell.north);
  \draw[arr] (shell.south -| ober) -- (ober.north);
  \draw[arr] (shell.south -| plus) -- (plus.north);
  \draw[arr2] (ober.south) -- (comp.north);
  \node[lbl, anchor=west] at (-3.42,-2.7)
    {setup and marks in, reports out};
  \draw[arr2] (plus.south) -- (eng.north);
  \node[lbl, anchor=west] at (3.78,-2.7)
    {alerts, flags, decisions, closure};
  \draw[arr] (comp.east) -- (eng.west);
  \node[font=\scriptsize, color=obskyd, anchor=south] at (0,-4.12)
    {$A_c^{(t)}$};
  \draw[arr2] (comp.south) -- (-3.6,-6.58);
  \node[lbl, anchor=west] at (-3.42,-5.75)
    {reads matrices and marks};
  \draw[arr2] (eng.south) -- (3.6,-6.58);
  \node[lbl, anchor=west] at (3.78,-5.75)
    {reads history, writes decisions and log};
\end{tikzpicture}}
\caption{Architecture of OBER+.}
\label{fig:arch}
\end{figure*}
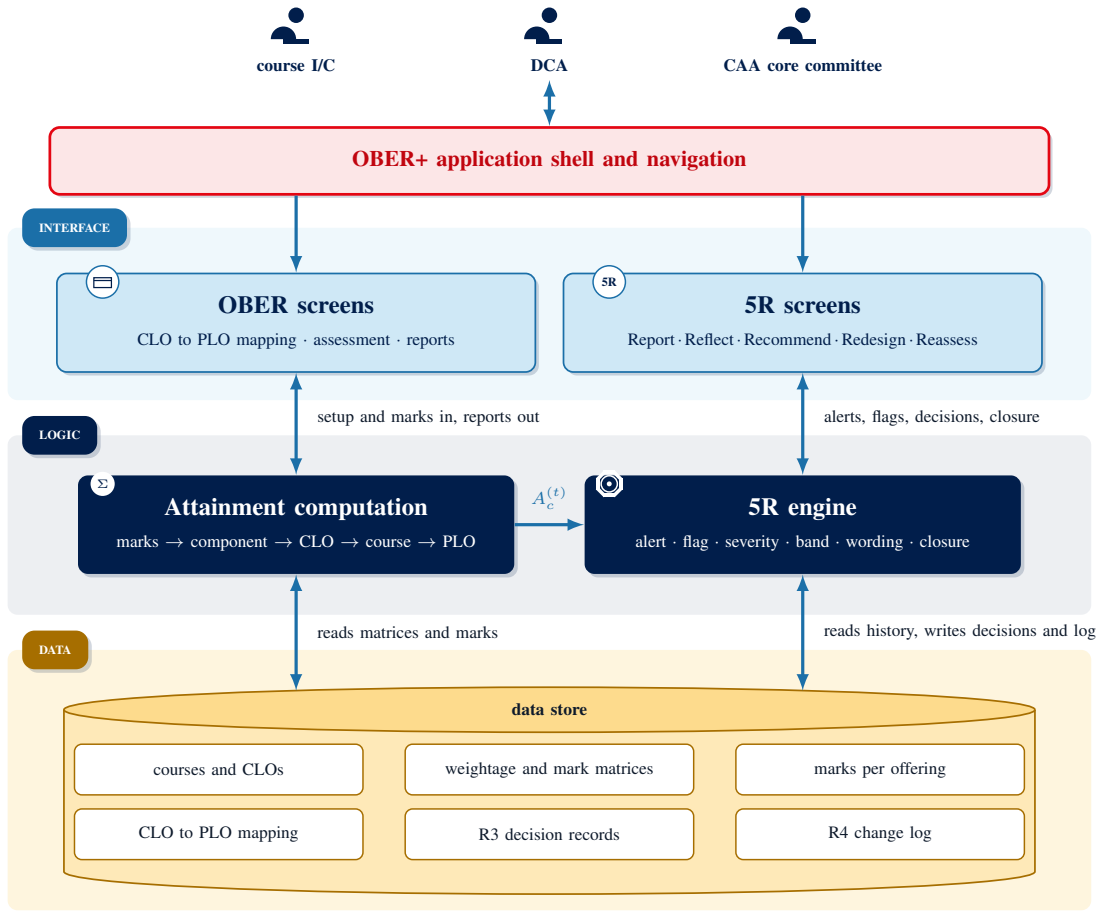

Fig.~\ref{fig:arch} shows the three layers. The interface layer carries
the reproduced reporting screens, covering CLO to PLO mapping,
assessment and reports, beside five screens, one per stage. The logic
layer holds the attainment computation of
(\ref{eq:component})--(\ref{eq:plo}) and a rule engine that computes
the alerts, the flags, the severity, the bands, the revision
classification and the gap closure. The data layer stores the courses
with their CLOs, the two distribution matrices, the marks per offering,
the CLO to PLO mappings, the decision records and the change log.

A single interface joins the two logic modules, across which the
computation supplies $A_c^{(t)}$ and nothing else. That narrowness is
what allows the rules to be adopted without the implementation, since
any institution whose own platform reports attainment per offering can
supply the same quantity.

\section{Evaluation Design}
\label{sec:evaldesign}

\subsection{Data Set}

The platform is an institutional system rather than a departmental one.
Every course in-charge of every programme on the campus enters
outcomes, evaluation components, mappings and marks into it for each
delivery, so its record spans the institution. The material used here
is a sample of that record, two courses in the Department of Computer
Science for which the author is the course in-charge and therefore able
to report the full computation chain, the outcome statements as issued,
and the figures behind them. What the sample bounds is the claims made
here, not the contents of the platform.

CS~F351 Theory of Computation is a third-year core course delivered
once a year, with 222 students enrolled in 2024-25 and 238 in 2025-26.
Its outcome statements are on record for three consecutive deliveries,
from the First Semester of 2023-24 to the First Semester of 2025-26,
each taken from the handout issued to students that year. Attainment is
on record for two of them, 2024-25 and 2025-26. The platform was
launched in the First Semester of 2025-26, so the 2024-25 attainment is
a historical record reconstructed from the departmental spreadsheet
that preceded it, which is the ordinary position of any institution
adopting such a platform. CS~F459 Computer Vision is an elective of 32
students with statements on record for 2025-26 and 2026-27 and
attainment for 2025-26. Cohort sizes therefore differ by roughly an
order of magnitude between the two courses, which makes a class mean a
noisier quantity in the elective than in the core course.

Each delivery contributes the outcome statements as issued in the
handout and entered in the platform, the evaluation components with
their weightage distribution matrix, the attainment reported for each
outcome and for the course, and the component-level attainment behind
that. Nothing finer grained than a class mean enters the analysis, so
no personally identifiable student data was processed. The attainment
target is 60 percent throughout. In total the record yields fifteen
statement pairs across three consecutive-delivery transitions and
fifteen outcome-level attainment figures across three deliveries, which
is enough to exercise every rule that two deliveries can reach and to
establish existence rather than frequency.

Three rules reach beyond two deliveries. The flag of (\ref{eq:flag})
needs an outcome to be below target in at least two of its last three
deliveries, the severity band of (\ref{eq:band}) applies only to a
flagged item, and the gap closure of (\ref{eq:closure}) needs a
delivery on either side of a logged change. These are demonstrated on the constructed record of
Section~\ref{sec:constructed}.

\subsection{Verification Procedure}

Verification of the attainment computation compares two independently
obtained figures for every outcome of both courses. The first is the
figure the platform displays. The second applies the documented rule of
(\ref{eq:clo}) to the component-level attainment the platform publishes
and to the weightage distribution matrix printed in the course handout.
Component figures are published to two decimal places, giving a
recomputed figure a rounding uncertainty near half a percentage point,
so agreement is defined throughout as a difference below one percentage
point. Where the two figures disagree, an alternative weighting drawn
from the mark distribution matrix is tested against the reported
figure, which identifies the rule the platform applies in place of the
documented one.

\subsection{Revision Detection Procedure}

The rule of (\ref{eq:revision}) produces a judgement rather than an
arithmetic result, and is examined on two bodies of material.

The first is a set of twenty-four pairs of outcome statements
constructed from the outcomes of the two courses to cover the four
classes of (\ref{eq:revision}), each labelled before the rule was run.
Because the threshold $\tau = 0.90$ was set on that same set,
agreement on it is a measure of internal consistency and not an
out-of-sample result.

The second is every pair of statements sharing an outcome number across
three consecutive-delivery transitions, fifteen pairs drawn from the
handouts issued for CS~F351 in 2023-24, 2024-25 and 2025-26 and for
CS~F459 in 2025-26 and 2026-27. Teaching teams wrote these
statements in the ordinary course of revising a handout, and none was
consulted while the threshold was set. None was labelled either, so the
returns on this material are characterised by three properties that
follow from the similarities alone, namely the position of the
threshold within the observed distribution, the sensitivity of the
classifications to that threshold, and the outcome of the alignment
check of (\ref{eq:align}). No accuracy figure is reported on real data.

\section{Results}
\label{sec:findings}

\subsection{Outcome Redefinition Across Deliveries}

Table~\ref{tab:statements} gives the five outcomes of CS~F351 as issued
in each of two consecutive deliveries. They are not revisions of one
another in any ordinary sense. The 2024-25 set runs from a foundational
grasp of automata theory through closure properties and equivalences,
parse trees and ambiguity, and Turing machines and undecidability, to
the complexity classes P and NP. The 2025-26 set runs from alphabets,
strings and proof techniques through finite automata and regular
expressions, context-free grammars and pushdown automata, to Turing
machines with decidability and complexity merged into one outcome.

Table~\ref{tab:revision} gives the classification of the five
same-numbered pairs, which is the comparison an attainment series makes
implicitly. None returns Unchanged. Two return Replacement and two a
change of cognitive level, while the pair numbered two returns
Paraphrase, for reasons Section~\ref{sec:fragile} sets out.

Comparing each 2025-26 statement against all five 2024-25 statements,
rather than against the one sharing its number, locates where the
subject matter went. The 2025-26 outcome numbered five, on Turing
machines, decidability and complexity, lies closest to the 2024-25
outcome numbered \emph{four}, on Turing machines and undecidability, at
similarity 0.955 against 0.933 for the outcome sharing its number. The
2025-26 outcome numbered four, on pushdown automata, reaches only 0.871
against any 2024-25 statement and has no counterpart in the earlier
set. Subject matter moved between outcome numbers, and one outcome is
new.

\begin{figure}[!t]
\centering
\begin{tikzpicture}
\begin{axis}[
  name=A,
  width=0.56\columnwidth, height=44mm,
  ybar, bar width=3.4pt,
  ymin=0, ymax=80, ytick={0,20,40,60,80},
  ylabel={Attainment (\%)},
  symbolic x coords={CLO1,CLO2,CLO3,CLO4,CLO5},
  xtick=data, xticklabel style={rotate=90, font=\tiny},
  enlarge x limits=0.13,
  ymajorgrids, grid style={obsky!25}, axis line style={obmuted},
  tick label style={font=\tiny, color=obink},
  label style={font=\scriptsize, color=obink},
  title={\scriptsize (a) CS~F351}, title style={yshift=-2pt},
]
\addplot[fill=obamber, draw=obamberd, line width=0.4pt] coordinates {
  (CLO1,66) (CLO2,67) (CLO3,70) (CLO4,65) (CLO5,60)};
\addplot[fill=obred, draw=obredd, line width=0.4pt] coordinates {
  (CLO1,53) (CLO2,30) (CLO3,52) (CLO4,42) (CLO5,38)};
\draw[dashed, obnavy, line width=0.8pt]
  (rel axis cs:0,0.75) -- (rel axis cs:1,0.75);
\end{axis}
\begin{axis}[
  name=B, at={(A.east)}, xshift=9mm, anchor=west,
  width=0.56\columnwidth, height=44mm,
  ybar, bar width=3.4pt,
  ymin=0, ymax=80, ytick={0,20,40,60,80}, yticklabels={,,},
  symbolic x coords={CLO1,CLO2,CLO3,CLO4,CLO5},
  xtick=data, xticklabel style={rotate=90, font=\tiny},
  enlarge x limits=0.13,
  ymajorgrids, grid style={obsky!25}, axis line style={obmuted},
  tick label style={font=\tiny, color=obink},
  title={\scriptsize (b) CS~F459}, title style={yshift=-2pt},
]
\addplot[fill=obsky, draw=obskyd, line width=0.4pt] coordinates {
  (CLO1,39) (CLO2,59) (CLO3,54) (CLO4,50) (CLO5,49)};
\draw[dashed, obnavy, line width=0.8pt]
  (rel axis cs:0,0.75) -- (rel axis cs:1,0.75)
  node[pos=1, anchor=south east, font=\tiny, color=obnavy] {Target 60};
\end{axis}
\node[font=\tiny, color=obink, anchor=north]
  at ($(A.south)!0.5!(B.south)$) [yshift=-7mm]
  {\textcolor{obamber}{$\blacksquare$}\,CS~F351 2024-25 \quad
   \textcolor{obred}{$\blacksquare$}\,CS~F351 2025-26 \quad
   \textcolor{obsky}{$\blacksquare$}\,CS~F459 2025-26};
\end{tikzpicture}
\caption{Outcome attainment in the live record.}
\label{fig:real}
\end{figure}
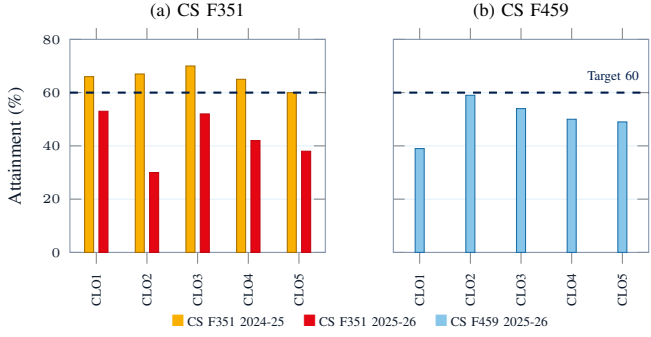

Table~\ref{tab:attain} gives outcome attainment for both deliveries
against the 60 percent target, plotted for both courses in
Fig.~\ref{fig:real}. Read as a series, the course falls from 65.0 to
40.3 percent and every outcome falls, by 13, 37, 18, 23 and 22
percentage points. Read beside Table~\ref{tab:revision}, that series
compares quantities which do not refer to the same learning, and the
23-point fall recorded against the fourth outcome sets attainment on
Turing machines against attainment on pushdown automata. The evaluation
scheme changed alongside the outcomes, from five components to six, and
the mapping to programme outcomes changed with them, so three sources
of incomparability coincide at a single boundary between consecutive
deliveries of one core course. None of the three leaves any trace in
the attainment record as current practice keeps it.

What survives the boundary is the alert of (\ref{eq:alert}), which
compares an outcome with its target inside one delivery. No outcome of
CS~F351 falls below target in 2024-25. All five do in 2025-26, as do
all five of CS~F459. Ten of the fifteen outcome figures in the live
record lie below target, and each raises an alert in the delivery where
it occurred.

\subsection{Attainment Computation Verification}

Table~\ref{tab:verify} sets the attainment the platform reports for
each outcome beside the figure its documented rule produces. Four of
the ten agree within the half-point rounding uncertainty. Six do not,
four of those by more than four percentage points, the widest being the
fifth outcome of CS~F351 at 38.0 percent reported against 46.6 percent
computed.

Weighting each component by its share of the marks allocated to the
outcome, which is the mark distribution matrix rather than the
weightage distribution matrix, reproduces all five reported figures for
CS~F351 to within 0.53 of a percentage point, and reproduces the course
figure at 40.13 percent against 40.28 reported, where the documented
rule gives 41.59. The two matrices coincide only where a component's
share of an outcome's marks equals its weightage, which is why four
figures agree and six do not. The platform's authors have been informed
and a correction is scheduled.

Two consequences reach beyond this campus. The defect is invisible to
users, since the figures are plausible, internally consistent, reported
to two decimal places, and produced by screens that show no arithmetic.
It surfaced only under an independent recomputation against a published
matrix. The design presented here is unaffected, because every rule
reads whatever attainment the platform reports and none recomputes it,
so the rules ran correctly on figures the platform had wrong. That
property is deliberate and was tested here for the first time. It is
also a limitation, since a framework that reads a number without
checking it will report a shortfall partly attributable to the reading.
The general principle is that traceability is only as sound as the
evidence entering the trace, so verification of the attainment value
belongs ahead of the cycle rather than beside it.

\subsection{Outcome Revision Detection}
\label{sec:fragile}

Over the fifteen pairs the rule returns five Unchanged, five Paraphrase,
three Replacement and two changes of cognitive level. The five
Unchanged are pairs identical character for character, correct by
construction and carrying no information about the rule, which leaves
ten judgements. Since none was labelled, these returns are
characterised below by three properties of the similarities of
(\ref{eq:sim}) rather than assessed for correctness.

\subsubsection{Threshold Position}

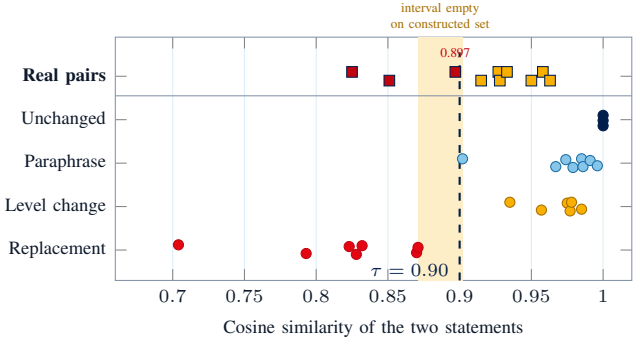
\begin{figure}[!t]
\centering
\begin{tikzpicture}
\begin{axis}[
  width=0.95\columnwidth, height=48mm,
  xmin=0.66, xmax=1.02, ymin=0.3, ymax=5.9,
  ytick={1,2,3,4,5},
  yticklabels={Replacement, {Level change}, Paraphrase, Unchanged,
               {\bfseries Real pairs}},
  yticklabel style={align=right, font=\scriptsize},
  xlabel={Cosine similarity of the two statements},
  axis line style={obmuted}, xmajorgrids, grid style={obsky!30},
  tick label style={font=\scriptsize, color=obink},
  label style={font=\scriptsize, color=obink}, clip=false,
]
\fill[obamber!22] (axis cs:0.871,0.3) rectangle (axis cs:0.902,5.9);
\node[font=\tiny, color=obamberd, anchor=south, align=center]
  at (axis cs:0.8865,5.9) {interval empty\\ on constructed set};
\draw[obnavy, line width=0.8pt, dashed] (axis cs:0.90,0.3) -- (axis cs:0.90,5.55);
\node[font=\scriptsize, color=obnavy, anchor=east] at (axis cs:0.899,0.55) {$\tau = 0.90$};
\draw[obmuted, line width=0.4pt] (axis cs:0.66,4.55) -- (axis cs:1.02,4.55);
\addplot[only marks, mark=*, mark size=1.9pt, draw=obredd, fill=obred]
  coordinates {(0.704,1.12) (0.793,0.92) (0.823,1.08) (0.828,0.90)
               (0.832,1.10) (0.870,0.94) (0.871,1.06)};
\addplot[only marks, mark=*, mark size=1.9pt, draw=obamberd, fill=obamber]
  coordinates {(0.935,2.10) (0.957,1.92) (0.975,2.08) (0.977,1.90)
               (0.978,2.10) (0.985,1.94)};
\addplot[only marks, mark=*, mark size=1.9pt, draw=obskyd, fill=obsky]
  coordinates {(0.902,3.10) (0.967,2.92) (0.974,3.08) (0.979,2.90)
               (0.985,3.10) (0.986,2.92) (0.991,3.06) (0.996,2.94)};
\addplot[only marks, mark=*, mark size=1.9pt, draw=obnavy, fill=obnavy]
  coordinates {(1.000,4.10) (1.000,3.98) (1.000,3.86)};
\addplot[only marks, mark=square*, mark size=2.0pt, draw=obnavy, fill=obredd]
  coordinates {(0.825,5.10) (0.851,4.90) (0.897,5.10)};
\addplot[only marks, mark=square*, mark size=2.0pt, draw=obnavy, fill=obamber]
  coordinates {(0.915,4.90) (0.927,5.10) (0.928,4.90) (0.933,5.10)
               (0.950,4.90) (0.958,5.10) (0.963,4.90)};
\node[font=\tiny, color=obredd, anchor=south] at (axis cs:0.897,5.22) {0.897};
\end{axis}
\end{tikzpicture}
\caption{Real pairs against the constructed distribution.}
\label{fig:classifier}
\end{figure}

Similarities on the twenty-four constructed pairs separate into two
groups with an interval of 0.031 between them, from 0.871 to 0.902,
containing no pair, and the threshold sits inside it. The ten real
non-identical pairs give no such separation. Their similarities, in
ascending order, are 0.825, 0.851, 0.897, 0.915, 0.927, 0.928, 0.933,
0.950, 0.958 and 0.963, the widest gap anywhere in that sequence being
0.046 between 0.851 and 0.897. The value closest to the threshold from
below, 0.897, falls inside the interval the constructed set left empty,
and the nearest value above it, 0.915, lies just beyond that interval,
as Fig.~\ref{fig:classifier} shows for both distributions on one axis.

The consequence for the classifications is narrower than that continuum
suggests. Moving the threshold by 0.005 changes none of the fifteen
classifications when raised and one when lowered, moving it by 0.02
changes one in either direction, and moving it by 0.05 changes five or
two according to direction. Within any plausible range the rule is
therefore stable, although the pair at 0.897 is decided by three
thousandths. What the constructed evaluation
showed was a boundary standing clear of the data, and the real data
does not reproduce that separation.

\subsubsection{Numbering Alignment}

The check of (\ref{eq:align}) requires neither the threshold nor the
taxonomy level, asking only which outcome of the previous delivery each
statement lies nearest to, and whether that is the outcome carrying the
same number.

Table~\ref{tab:align} gives the answer for all fifteen pairs, nine of
which have the same-numbered outcome as their nearest match. Alignment
is trivially perfect in the transition whose statements are identical,
so the informative figure covers the ten pairs that are not identical
character for character, and there the numbering follows the subject
matter in four cases and fails in six.

Several failures are wide rather than marginal. The fifth outcome of
CS~F351 in 2025-26 lies nearest to the fourth outcome of 2024-25 at
0.955 against 0.933 for the outcome sharing its number. The third
outcome of CS~F459 in 2026-27 lies nearest to the fifth of 2025-26 at
0.910 against 0.851. The fourth outcome of CS~F459 in 2026-27 lies
nearest to the second of 2025-26 at 0.952 against 0.927. An attainment
series indexed by outcome number sets two different subjects against
each other in each case and reports the difference as movement.

The result rests on no threshold, no taxonomy level, no labelling and
no judgement of what counts as a paraphrase. An ordering of cosine
similarities says that in six of the ten outcomes whose wording changed
at all, the number no longer attaches to the subject matter it carried
in the previous delivery. Any institution reading outcome attainment as a
series indexed by outcome number is exposed to that, and no system in
Section~\ref{sec:related} tests for it.

Two observations bear on where the rule understates change, and both
are observations rather than findings, since no labelling was done. The
five Paraphrase returns concentrate in one course, four of the five
outcomes of CS~F459 across the 2025-26 to 2026-27 transition, at
similarities between 0.927 and 0.963. That same transition contains
three of the six misalignments, which is not what a set of reworded
statements would produce. The pair returned as a Paraphrase at 0.928
sets ``Apply feature detection, description, and segmentation
techniques for image representation and region analysis'' against
``Apply machine learning techniques for image classification, including
linear classifiers, loss functions, optimization, and neural networks
with backpropagation'', two statements written in the vocabulary of
computer vision and both beginning with Apply, which is the
configuration in which an averaged static word vector and a
leading-verb lookup have least purchase.

The second observation concerns the levelling. The campus verb list,
which follows \cite{anderson}, assigns Develop and Construct to both
Applying and Creating and Explain to both Understanding and Evaluating,
and all three lead outcomes in these handouts. Where level sets
intersect no change of level can be asserted, so the rule falls back to
Paraphrase whenever similarity exceeds the threshold, which is what
happens in the pair just described. A levelling step reading only the
first word cannot be deterministic against a list that is itself
ambiguous, and an arbitrary choice between the two levels would make
the classification depend on an undocumented tie-break. A Paraphrase
from this rule is therefore unconfirmed rather than a finding of no
change, and establishing at what rate such returns are wrong requires
the labelling described in Section~\ref{sec:future}.

\subsection{Framework Illustration on a Constructed Record}
\label{sec:constructed}

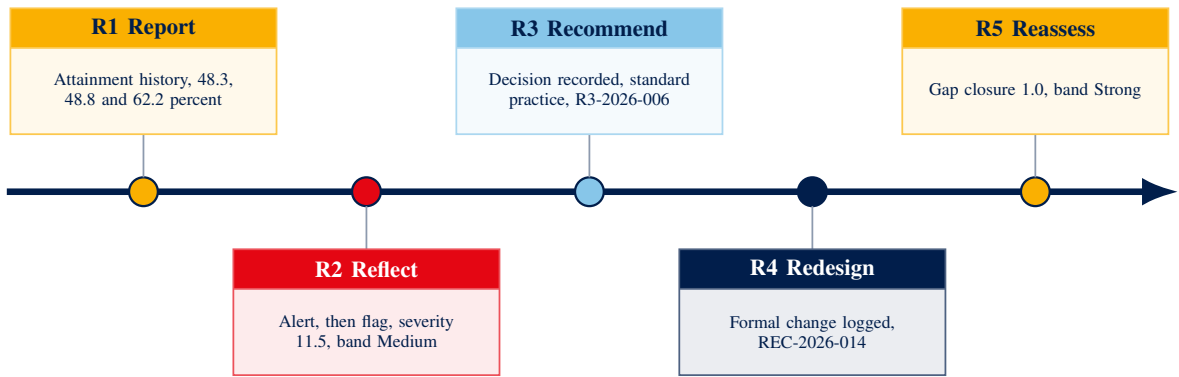
\begin{figure*}[!t]
\centering
\resizebox{0.86\textwidth}{!}{%
\begin{tikzpicture}[>=Latex]
\draw[obnavy,line width=2.6pt,->] (0,0) -- (16.3,0);
\foreach \x/\c in {1.9/obamber, 5.0/obred, 8.1/obsky, 11.2/obnavy, 14.3/obamber}{
  \fill[\c] (\x,0) circle (0.20);
  \draw[obnavy,line width=0.9pt] (\x,0) circle (0.20);}
\draw[obnavy!45,line width=0.7pt] (1.9,0.20)  -- (1.9,0.80);
\draw[obnavy!45,line width=0.7pt] (8.1,0.20)  -- (8.1,0.80);
\draw[obnavy!45,line width=0.7pt] (14.3,0.20) -- (14.3,0.80);
\draw[obnavy!45,line width=0.7pt] (5.0,-0.20)  -- (5.0,-0.80);
\draw[obnavy!45,line width=0.7pt] (11.2,-0.20) -- (11.2,-0.80);
\trstage{1.9}{0.80}{2.55}{obamber}{obnavy}{R1 Report}
  {Attainment history, 48.3, 48.8 and 62.2 percent}
\trstage{8.1}{0.80}{2.55}{obsky}{obnavy}{R3 Recommend}
  {Decision recorded, standard practice, R3-2026-006}
\trstage{14.3}{0.80}{2.55}{obamber}{obnavy}{R5 Reassess}
  {Gap closure 1.0, band Strong}
\trstage{5.0}{-2.55}{-0.80}{obred}{white}{R2 Reflect}
  {Alert, then flag, severity 11.5, band Medium}
\trstage{11.2}{-2.55}{-0.80}{obnavy}{white}{R4 Redesign}
  {Formal change logged, REC-2026-014}
\end{tikzpicture}}
\caption{One CLO traced from detection to the following delivery.}
\label{fig:trace}
\end{figure*}

The flag, the severity band and the gap closure need three deliveries
where the live record holds two, and are illustrated in
Table~\ref{tab:constructed} on a small constructed record carrying no
claim made in this paper. The illustration holds an outcome failing in
two deliveries of three, flagged with an average shortfall of 11.5
points by (\ref{eq:severity}) and a Medium band by (\ref{eq:rho}), a
decision recorded as (\ref{eq:decision}) against that flag, a change
logged as (\ref{eq:logentry}) implementing it, and a gap closure of 1.0
banded Strong in the delivery that follows. It also holds an
intervention that did not work, at closure 0.11 banded Very limited,
and an outcome flagged and left unaddressed, so that inaction and
failure appear as plainly as success. Fig.~\ref{fig:trace} follows the
first of those outcomes from detection to the delivery that follows the
change. The illustration demonstrates that the rules compute and that
the identifiers connect a shortfall to a decision, a change and the
movement recorded afterwards, and it demonstrates nothing about
courses.

\begin{table*}[!t]
\caption{Outcome statements of CS~F351 as issued in each delivery.}
\label{tab:statements}
\centering
\footnotesize
\renewcommand{\arraystretch}{1.25}
\begin{tabular}{L{9mm}|L{77mm}|L{77mm}}
\hline
\textbf{} & \textbf{First Semester 2024-25} & \textbf{First Semester 2025-26} \\
\hline
CLO1 & Develop a foundational grasp of automata theory and formal
languages, including their practical implications in computer science
contexts. & Explain the fundamental concepts of alphabets, strings,
languages, infinite sets, closure properties, and proof techniques. \\
\hline
CLO2 & Apply closure properties to various language classes, explore
equivalences between different formal models, and evaluate the
consequences for language manipulation. & Construct finite automata and
regular expressions for language specification, including closure
properties and equivalence. \\
\hline
CLO3 & Demonstrate skills in creating parse trees, resolving ambiguity
in context-free grammars, and utilizing parsing techniques for
effective language structure analysis. & Design context-free grammars
by applying parsing strategies and addressing ambiguity in language
representation. \\
\hline
CLO4 & Analyze Turing machines as abstract computational tools,
differentiating solvable problems and exploring the concept of
undecidability, particularly in relation to the halting problem. &
Develop pushdown automata models for context-free language recognition
and analysis. \\
\hline
CLO5 & Examine complexity classes P and NP, differentiate their
attributes, comprehend polynomial-time reductions, and appreciate
NP-completeness' significance in computational complexity theory. &
Develop Turing Machine models and analyze computational problems in
terms of decidability and complexity, including recursive languages,
the halting problem, and NP-completeness. \\
\hline
\end{tabular}
\end{table*}

\begin{table}[!t]
\caption{Classification of the five real revisions of CS~F351.}
\label{tab:revision}
\centering
\scriptsize
\renewcommand{\arraystretch}{1.2}
\begin{tabular}{l|c|L{20mm}|L{20mm}|l}
\hline
\textbf{Pair} & \textbf{$s_t$} & \textbf{Leading verb 2024-25} &
\textbf{Leading verb 2025-26} & \textbf{Class} \\
\hline
CLO1 & 0.897 & Develop \newline Apply or Create &
Explain \newline Understand or Evaluate & Replacement \\
\hline
CLO2 & 0.915 & Apply \newline Apply &
Construct \newline Apply or Create & Paraphrase \\
\hline
CLO3 & 0.950 & Demonstrate \newline Understand &
Design \newline Create & Level change \\
\hline
CLO4 & 0.825 & Analyze \newline Analyze &
Develop \newline Apply or Create & Replacement \\
\hline
CLO5 & 0.933 & Examine \newline Analyze &
Develop \newline Apply or Create & Level change \\
\hline
\end{tabular}
\end{table}

\begin{table}[!t]
\caption{Outcome attainment in the live record, against a target of 60.}
\label{tab:attain}
\centering
\footnotesize
\renewcommand{\arraystretch}{1.2}
\begin{tabular}{l|c|c|c}
\hline
& \multicolumn{2}{c|}{\textbf{CS~F351}} & \textbf{CS~F459} \\
\cline{2-4}
\textbf{Outcome} & \textbf{2024-25} & \textbf{2025-26} &
\textbf{2025-26} \\
\hline
CLO1   & 66.0 & 53.0 & 39.0 \\
CLO2   & 67.0 & 30.0 & 59.0 \\
CLO3   & 70.0 & 52.0 & 54.0 \\
CLO4   & 65.0 & 42.0 & 50.0 \\
CLO5   & 60.0 & 38.0 & 49.0 \\
\hline
Course & 65.0 & 40.3 & 50.6 \\
\hline
Alerts raised & 0 of 5 & 5 of 5 & 5 of 5 \\
\hline
Enrolled students & 222 & 238 & 32 \\
\hline
\end{tabular}
\end{table}

\begin{table*}[!t]
\caption{Attainment reported by the platform against its documented computation.}
\label{tab:verify}
\centering
\footnotesize
\renewcommand{\arraystretch}{1.2}
\begin{tabular}{l|l|c|c|c|c|c}
\hline
\textbf{Course} & \textbf{Outcome} &
\textbf{Reported by} & \textbf{Documented rule} &
\textbf{Difference} & \textbf{Agrees within} &
\textbf{Weighting by} \\
 & & \textbf{platform (\%)} & \textbf{(\ref{eq:clo}) (\%)} &
\textbf{(pts)} & \textbf{rounding} & \textbf{mark share (\%)} \\
\hline
CS~F351 & CLO1 & 53.0 & 50.2 & $+2.8$ & No  & 52.5 \\
        & CLO2 & 30.0 & 28.2 & $+1.8$ & No  & 29.9 \\
        & CLO3 & 52.0 & 59.2 & $-7.2$ & No  & 51.9 \\
        & CLO4 & 42.0 & 42.8 & $-0.8$ & Yes & 42.5 \\
        & CLO5 & 38.0 & 46.6 & $-8.6$ & No  & 37.6 \\
\cline{2-7}
        & Course & 40.28 & 41.59 & $-1.31$ & No & 40.13 \\
\hline
CS~F459 & CLO1 & 39.0 & 39.5 & $-0.5$ & Yes & not available \\
        & CLO2 & 59.0 & 63.0 & $-4.0$ & No  & not available \\
        & CLO3 & 54.0 & 54.0 & $\pm0.0$ & Yes & not available \\
        & CLO4 & 50.0 & 55.2 & $-5.2$ & No  & not available \\
        & CLO5 & 49.0 & 49.6 & $-0.6$ & Yes & not available \\
\hline
\end{tabular}
\end{table*}

\begin{table}[!t]
\caption{Nearest earlier outcome for each outcome of each delivery.}
\label{tab:align}
\centering
\scriptsize
\renewcommand{\arraystretch}{1.15}
\begin{tabular}{l|c|c|c|c}
\hline
\textbf{Delivery and} & \textbf{Same} & \textbf{Nearest} & \textbf{Its} &
\textbf{Aligned} \\
\textbf{outcome} & \textbf{number} & \textbf{number} & \textbf{value} & \\
\hline
\multicolumn{5}{l}{\textit{CS~F351, 2023-24 to 2024-25}} \\
\hline
CLO1 & 1.000 & CLO1 & 1.000 & Yes \\
CLO2 & 1.000 & CLO2 & 1.000 & Yes \\
CLO3 & 1.000 & CLO3 & 1.000 & Yes \\
CLO4 & 1.000 & CLO4 & 1.000 & Yes \\
CLO5 & 1.000 & CLO5 & 1.000 & Yes \\
\hline
\multicolumn{5}{l}{\textit{CS~F351, 2024-25 to 2025-26}} \\
\hline
CLO1 & 0.897 & CLO2 & 0.913 & No \\
CLO2 & 0.915 & CLO2 & 0.915 & Yes \\
CLO3 & 0.950 & CLO3 & 0.950 & Yes \\
CLO4 & 0.825 & CLO1 & 0.871 & No \\
CLO5 & 0.933 & CLO4 & 0.955 & No \\
\hline
\multicolumn{5}{l}{\textit{CS~F459, 2025-26 to 2026-27}} \\
\hline
CLO1 & 0.958 & CLO2 & 0.970 & No \\
CLO2 & 0.928 & CLO2 & 0.928 & Yes \\
CLO3 & 0.851 & CLO5 & 0.910 & No \\
CLO4 & 0.927 & CLO2 & 0.952 & No \\
CLO5 & 0.963 & CLO5 & 0.963 & Yes \\
\hline
\end{tabular}
\end{table}

\begin{table*}[!t]
\caption{Illustration of the three rules the live record cannot reach.}
\label{tab:constructed}
\centering
\footnotesize
\renewcommand{\arraystretch}{1.2}
\begin{tabular}{L{15mm}|L{28mm}|L{15mm}|L{40mm}|L{34mm}|L{22mm}}
\hline
\textbf{Item} & \textbf{Attainment over three deliveries (\%)} &
\textbf{Flag and band} & \textbf{Decision recorded} &
\textbf{Change logged} & \textbf{Gap closure} \\
\hline
Course A, CLO4 & 48.3, 48.8, 62.2 & Yes, shortfall 11.5, Medium &
R3-2026-006, standard practice, worked examples &
REC-2026-014, formal, implements R3-2026-006 & 1.0, Strong \\
\hline
Course A, CLO3 & 62.3, 57.0, 60.6 & No & None &
REC-2026-015, detected by the wording rule at a change of course I/C &
1.0, Strong \\
\hline
Course B, CLO4 & below target throughout & Yes, shortfall 41.4, Low &
R3-2026-011, innovative practice, industry-sourced content &
REC-2026-021, formal, implements R3-2026-011 & 0.11, Very limited \\
\hline
Course B, CLO3 & below target throughout & Yes, Medium & None & None &
Not evaluable \\
\hline
\end{tabular}
\end{table*}

\section{Discussion}
\label{sec:discussion}

\subsection{Interpretation}

A framework built to read attainment across deliveries established, on
the first record it was given, that those deliveries must not be
compared. Every outcome of CS~F351 was redefined between 2024-25 and
2025-26, subject matter moved between outcome numbers, one outcome was
new, the evaluation scheme grew from five components to six and the
mapping to programme outcomes changed as well. A record reporting the
course as falling from 65.0 to 40.3 percent would have been read as a
collapse in student attainment, and would have prompted corrective
action against a decline the record cannot support.

Precision matters in stating that. Some part of a 25-point movement may
well be real, and the design cannot decompose it. The claim is that the
record cannot separate a fall in attainment from a redefinition of what
was measured, and that no system in Section~\ref{sec:related} would
have registered the difference. A decision taken on such a record rests
on a comparison that does not hold, which is a concrete instance of the
failure the curriculum analytics reviews describe when they report an
absence of evidence connecting tools to decisions and outcomes
\cite{drugova2024,desilva2025}.

Verification compounds the point. Six of ten outcome figures disagree
with the platform's own documented computation, and a different
weighting reproduces the disagreement exactly, so the figures a course
team would have acted on are not the figures the institution's
specification defines. Neither the redefinition nor the computation
defect is visible on any screen the platform presents. Both surfaced
because OBER+ subjected records the reporting system treats as settled
to a further check, of continuity in the one case and of the
computation itself in the other.

\subsection{Implications for Practice}

A course in-charge gains a record in place of a memory. Courses change
hands between deliveries, and what a predecessor tried, and whether it
worked, currently travels by conversation. Under the design the
attainment history, the signals raised on it, the decision taken and
the measured movement all travel with the course. A department gains a
basis for attention that does not rest on impression, since severity
bands and the distribution of closures identify which courses carry the
most severe flags, which flags carry no decision and which changes
produced little movement. An institution gains the account that
continuous improvement review asks for, held before the review rather
than assembled for it, and gains it without a new committee and without
moving the improvement decision away from the course in-charge.

The results add an implication the design did not anticipate. An
attainment series should not be published across a delivery boundary
until the outcomes on either side have been compared, and that
comparison has to be automatic, since no institution performs it by
hand across every course every year. Institutions reporting attainment
as a trend, which is most of those operating under the frameworks of
Section~\ref{sec:related}, may be reporting movements that partly
reflect redefinition. Checking costs one comparison per outcome per
delivery, and the check of (\ref{eq:align}) needs neither a threshold
nor a label.

Transfer follows from the interface rather than from the code, because
the rules read $A_c^{(t)}$ and nothing else. Any platform reporting
attainment per delivery supplies that quantity.

\subsection{Limitations}

Nothing reported here is evidence about learning. The design neither
teaches nor interacts with students, and no result bears on whether any
student learned more.

The record is small, covering two courses and three deliveries in one
department of one campus, which supports existence claims and no rate.

No real statement pair was labelled, so the returns of the
classification rule are reported without any assessment of whether they
are correct. A Paraphrase should be read as unconfirmed rather than as
a finding of no change. The alignment result is unaffected, depending
as it does on an ordering of similarities rather than on a
classification.

Real similarities form a continuum across a threshold that was set on
constructed pairs, which makes the rule an indicator that a comparison
needs review rather than a settled classification.

The flag, the severity band and the gap closure need three deliveries
and were exercised only on the constructed record of
Section~\ref{sec:constructed}.

The 2024-25 attainment predates the platform and was reconstructed from
a departmental spreadsheet, so it was not produced by the computation
that produced the 2025-26 figures, which is itself part of why the two
deliveries are not comparable.

The design finally reads whatever attainment the platform reports and
never recomputes it. That property kept the rules correct on defective
figures, and it also means a shortfall the framework reports may be
partly an artefact of the platform's computation.

\subsection{Future Work}
\label{sec:future}

The comparison that licenses the improvement cycle now precedes it in
priority, and the data for that comparison is already held. Since the
platform records the outcome statements of every course of every
programme for each delivery, the comparison run here on two courses can
be run across the institution by extracting statements per course per
delivery and applying the same rule. That would establish how often
outcomes are redefined without trace, in what way, whether redefinition
concentrates where a course changes hands, and whether it concentrates
in particular programmes. It requires no new instrument and no new
collection, only a query against a system already in operation, and it
would convert an existence result into a rate.

Evaluating the classification rule properly requires that material.
Labelling by at least two raters who taught the courses, with
inter-rater agreement reported and the threshold fixed on data disjoint
from the evaluation, would replace the observation reported here with a
measurement. A contextual encoder should be compared against the static
vectors used here under the same reproducibility constraint. The
levelling step needs replacing rather than tuning, since a classifier
trained on outcome statements \cite{li2022,gani2023,almatrafi2025}
levels a whole statement and is therefore not defeated by a verb that
the institutional list places at two levels, which is the failure that
produced the single missed redefinition reported above. Whether such a
classifier can be made reproducible enough for an accreditation record
is the open question, and it should be evaluated on that criterion
alongside accuracy.

The next stage is to evaluate the improvement cycle itself on the live
record rather than on a constructed one. A third delivery of CS~F351
will allow the flag, the severity band and the gap closure to run on
live data for the first time. Whether the mechanism
produces the benefit it is designed to produce would be established by
a study capable of supporting causal inference, whether through
staggered adoption across comparable deliveries or through the
comparison of courses that act upon a flag with those that defer a
response.

\section{Conclusion}

Institutions compute learning outcome attainment routinely and leave
the step from a weak result to an evaluated corrective action to
committees, which is why curriculum analytics reviews report an absence
of evidence about what such decisions achieve. This paper set out a
design that computes that step, as rules reading only the attainment a
local platform already reports, together with a rule that detects and
classifies revision of an outcome statement so that a series is never
compared across a redefinition. Applying it to the live record of two
real courses produced three results.

Every outcome of a core course was substantively redefined between
consecutive deliveries, subject matter moved between outcome numbers
and one outcome was new, while attainment continued to be reported
against unchanged numbers, so a record showing the course falling from
65.0 to 40.3 percent compares quantities that do not refer to the same
learning. Recomputing the platform's figures from its documented rule
showed six of ten differing by more than rounding explains, and
weighting components by their share of marks reproduced the pattern
exactly, which identified a defect since reported and scheduled for
correction. Across fifteen statement pairs from three
consecutive-delivery transitions, the outcome carrying a given number
was nearest to a differently numbered earlier outcome in six of the ten
pairs that were not identical character for character, which establishes without
threshold, taxonomy level or labelling that an attainment series
indexed by outcome number does not follow the subject matter it
reports.

Two of those three run against the design's own account of itself,
which is why they most need reporting. What the work establishes is
that attainment recorded across deliveries cannot be read as a series
until the outcomes on either side have been compared, that the
comparison is computable, and that an institution omitting it may act
on a movement it has not measured. Because that comparison can now be
verified, an institution can ask whether the corrective actions it
takes improve what its students learn, on figures that support the
question. Answering it is the next stage. The
implementation is public and the rules read a
single quantity, so any institution computing attainment for each
delivery of a course can run the same comparison on its own record.

\section*{Acknowledgment}
This work was developed under the LEAD Academics programme of BITS
Pilani, Leadership in Education, Advocacy and Policy-Development, and
the author thanks the programme for the structure within which it was
carried out.

\begin{IEEEbiography}[{\includegraphics[width=1in,height=1.25in,%
clip,keepaspectratio]{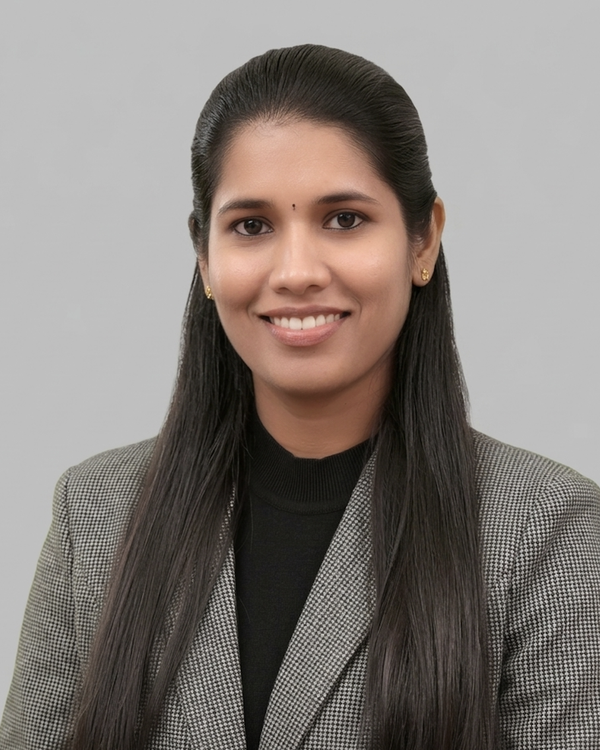}}]{Elakkiya Rajasekar}
received the B.E.\ degree in Computer Science and Engineering, the
M.E.\ degree in Software Engineering, and the Ph.D.\ degree, all from
Anna University, Chennai, India, in 2010, 2012, and 2018, respectively.
She is currently an Associate Professor with the Department of Computer
Science, Birla Institute of Technology and Science, Pilani, Dubai
Campus, Dubai International Academic City, Dubai, UAE, and serves as
Associate Head of the Anuradha and Prashanth Palakurthi Centre for
Artificial Intelligence Research (APPCAIR). She has published more than
95 research articles in IEEE Transactions, Elsevier, and Springer
venues, holds 6 patents, and has authored 12 books and 12 book
chapters. She has been recognised among the Top 2\% World Scientists by
the Stanford--Elsevier global rankings (2024--25 and 2025--26). She
serves as Chair of the ACM-W Asia Pacific Chapters, Chair of the ACM-W
Professional Dubai Chapter and Vice Chair of the ACM Dubai Professional
Chapter. Her research interests
include edge AI security, machine learning, adversarial robustness, and
sign language recognition.
\end{IEEEbiography}

\end{document}